\documentclass{article}
\usepackage{ijcai25}

\usepackage{times}
\usepackage{soul}
\usepackage{url}
\usepackage[hidelinks]{hyperref}
\usepackage[utf8]{inputenc}
\usepackage[small]{caption}
\usepackage{graphicx}
\usepackage{amsmath}
\usepackage{amsthm}
\usepackage{booktabs}
\usepackage{algorithm}
\usepackage{algorithmic}
\usepackage[switch]{lineno}

\usepackage{algorithm}
\usepackage{algorithmic}
\usepackage{amsmath}
\usepackage{bm}
\usepackage{multirow}
\usepackage{makecell}
\usepackage{subfigure}

\title{Logic Distillation: Learning from Code Function by Function\\ for Decision-making Tasks}

\author{
    Dong Chen\textsuperscript{\rm 1,2,3}, Shilin Zhang\textsuperscript{\rm 1}, Fei Gao\textsuperscript{\rm 1}, Yueting Zhuang\textsuperscript{\rm 4}, Siliang Tang\textsuperscript{\rm 4}, Qidong Liu\textsuperscript{\rm 1,2,3} \thanks{Corresponding Author}, Mingliang Xu\textsuperscript{\rm 1,2,3} \thanks{Corresponding Author}
    \affiliations
\textsuperscript{\rm 1}The School of Computer and Artificial Intelligence of Zhengzhou University\\
\textsuperscript{\rm 2}Engineering Research Center of Intelligent Swarm Systems, Ministry of Education\\
\textsuperscript{\rm3}National Supercomputing Center In Zhengzhou\\
\textsuperscript{\rm 4}Zhejiang University
    \emails
    \{chendongai,ieqdliu,iexumingliang\}@zzu.edu.cn, \{iszhangshilin1024,gaofei0191\}@gs.zzu.edu.cn,
    \{yzhuang,siliang\}@zju.edu.cn
}

\begin{document}
	
	\maketitle
	
	\begin{abstract}
		Large language models (LLMs) have garnered increasing attention owing to their powerful comprehension and generation capabilities. Generally, larger LLMs (L-LLMs) that require paid interfaces exhibit significantly superior performance compared to smaller LLMs (S-LLMs) that can be deployed on a variety of devices. Knowledge distillation (KD) aims to empower S-LLMs with the capabilities of L-LLMs, while S-LLMs merely mimic the outputs of L-LLMs, failing to get the powerful decision-making capability for new situations. Consequently, S-LLMs are helpless when it comes to continuous decision-making tasks that require logical reasoning. To tackle the identified challenges, we propose a novel framework called Logic Distillation (LD). Initially, LD employs L-LLMs to instantiate complex instructions into discrete functions and illustrates their usage to establish a function base. Subsequently, LD fine-tunes S-LLMs based on the function base to learn the logic employed by L-LLMs in decision-making. During testing, S-LLMs will yield decision-making outcomes, function by function, based on current states. Experiments demonstrate that with the assistance of LD, S-LLMs can achieve outstanding results in continuous decision-making tasks, comparable to, or even surpassing, those of L-LLMs. The code and data for the proposed method are provided for research purposes https://github.com/Anfeather/Logic-Distillation.
	\end{abstract}
	\section{Introduction}
	\label{Introduction}
	\begin{figure}[h]
		\centering
		\includegraphics[scale=0.55]{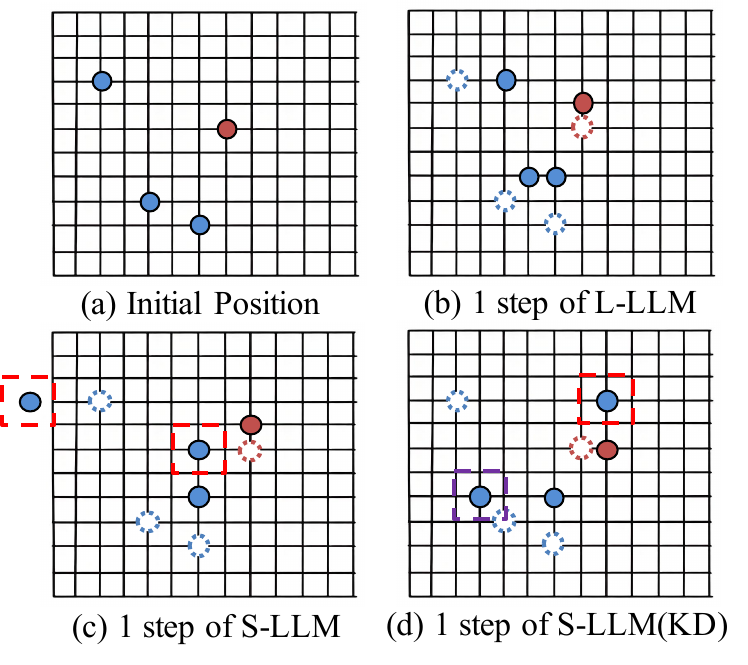}
		\caption{The outcome of one step in the pursuit game. }
		\label{Fig: motivation}
	\end{figure}
	
	Large language models (LLMs) \cite{ouyang2022training}, such as GPT-4 \cite{achiam2023gpt} and GLM-4 \cite{glm2024chatglm}, have been extensively applied owing to their powerful capabilities like comprehension and generation. Particularly, LLMs demonstrate superior performance in autonomous embodied agents, showcasing advanced decision-making capabilities grounded in comprehending instructions and logical reasoning \cite{xi2023rise}.

	Despite the remarkable capabilities of LLMs, such as GPT-4 and GLM-4, their substantial computational requirements render them impractical for deployment on most devices \cite{chen2024data,chen2024improving,chen2025kka}. On the other hand, numerous companies have attempted to develop relatively smaller open-source LLMs, including GLM-4-9B \cite{glm2024chatglm} and LLaMA-7B \cite{touvron2023llama}, which are compatible with consumer-grade GPUs like RTX 3090 Ti. In this paper, we refer to LLMs that cannot be deployed on most devices and require invocation through a paid interface as larger LLMs (L-LLMs), in contrast to smaller LLMs (S-LLMs) deployable on consumer-grade GPUs. Generally, L-LLMs exhibit significantly superior performance across various domains, particularly in logical reasoning. Nonetheless, S-LLMs have garnered extensive attention owing to their convenient deployment and cost-free nature. Consequently, an increasing number of researchers are focusing on Knowledge Distillation (KD) of LLMs \cite{gou2021knowledge}, where L-LLMs act as teachers imparting knowledge, while S-LLMs serve as students, mimicking the outputs of teachers \cite{xu2024survey}.
	

	While KD has been demonstrated to effectively enhance the capabilities of S-LLMs in numerous tasks \cite{dai2023auggpt}, endowing S-LLMs with the decision-making capability of L-LLMs continues to pose a significant challenge. 
	As shown in Figure \ref{Fig: motivation}, we present the decision-making outcome of a single step in a pursuit game, where LLMs control three blue dots chase an orange dot. Figure \ref{Fig: motivation}(a) displays the initial positions of four dots. Meanwhile, in Figure \ref{Fig: motivation}(b), the L-LLM (GLM-4) effectively comprehends the game rules, enabling informed decision-making based on the position of the orange dot. In contrast, Figure \ref{Fig: motivation}(c) reveals that the S-LLM (GLM-4-9B) lacks proficiency in following rules as the points enclosed by the red dashed line violate the rules of movement. Additionally, Figure \ref{Fig: motivation}(d) highlights the issue with KD, where the student merely mimics the output of the teacher without comprehending the logic behind the teacher’s decision-making. Specifically, in Figure \ref{Fig: motivation}(d), the orange dot moved one unit to the right, while the point enclosed by the purple dashed line moved one unit to the left. Such results stem from the KD process, where the S-LLM remembers the outputs the L-LLM has had in the past. 
	Based on the aforementioned analysis, we summarize the limitations of S-LLMs and KD in decision-making tasks as follows: 1. Ineffectiveness in following complex instructions: Despite extensive fine-tuning, S-LLMs continue to struggle with following intricate rules in decision-making tasks. 2. Failure to comprehend the logic of L-LLMs: Fine-tuned S-LLMs merely mimic the outputs of L-LLMs and lose their decision-making capabilities when encountering unknown scenarios.

	Recently, there has been considerable work aimed at establishing the translation between natural language and code, where a sentence is a logical code line \cite{feng2020codebert,chen2021evaluating}. Drawing inspiration from this, we suggest breaking down the decision-making logic of L-LLMs into multiple stages, with each stage being represented by a specific code function. Subsequently, S-LLMs engage in decision-making by learning and applying the pertinent functions. 
	More specifically, we propose Logic Distillation (LD). First, we leverage the powerful comprehension and logical reasoning capabilities of L-LLMs to decompose the rules governing decision-making tasks into multiple stages, forming different functions. Subsequently, we integrate these functions along with their comments, usage examples, etc., into a function base, and use it to fine-tune S-LLMs to enhance their ability to invoke relevant functions. Besides, to alleviate the issue of S-LLMs struggling to follow complex instructions, we propose transforming generation into selection. During each stage of decision-making, S-LLMs will select and invoke functions based on the current state.
	
\begin{figure}[t]
		\centering
		\includegraphics[scale=0.5]{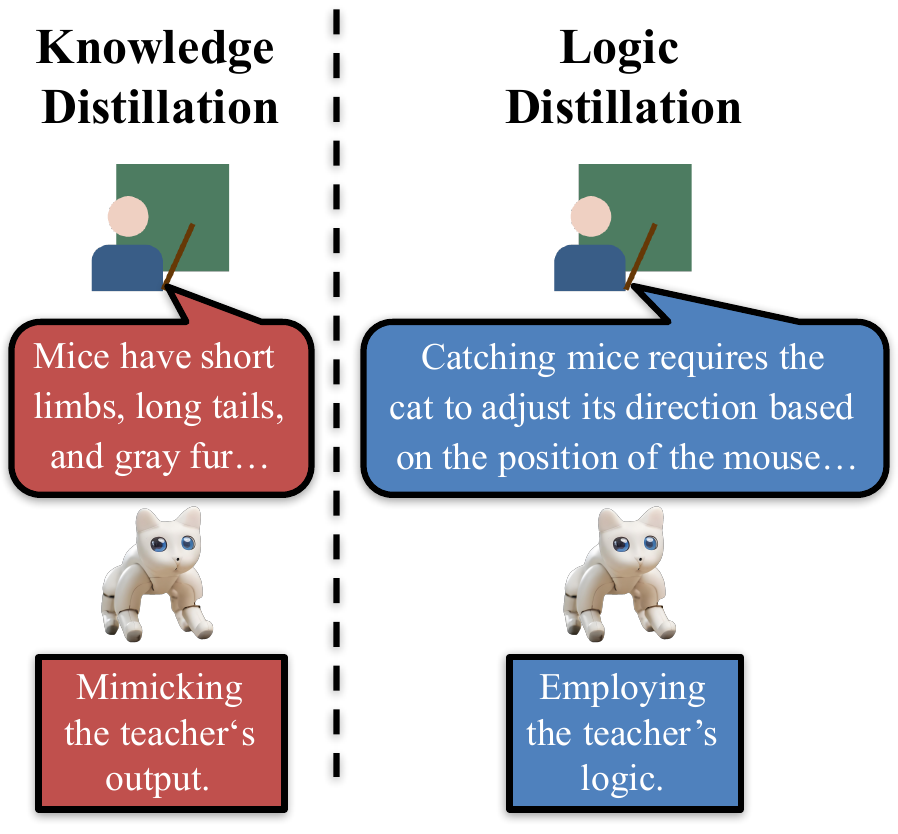}
		\caption{KD vs LD. KD aims to have smaller models mimic the output of larger models, while LD tries to enable smaller models to understand how larger models accomplish a task.}
		\label{Fig: LDvsKD}
	\end{figure}

	As depicted in Figure \ref{Fig: LDvsKD}, both KD and LD require guidance from a teacher with superior capabilities. KD emphasizes the imparting of content, which involves students mimicking the teacher's output. In contrast, the proposed LD concentrates on the underlying logic of task execution. The teacher decomposes a complex task into multiple basic logics, represented by functions, enabling the student to make decision function by function.
	
	The main contributions of this paper can be summarized as follows:
	\begin{itemize}
		\item We analyzed the issues of Knowledge Distillation in the context of decision-making in the LLMs era.
		\item We propose Logic Distillation to enable S-LLMs to accomplish decision-making tasks akin to L-LLMs.
		\item We conducted experiments in different scenarios to validate the effectiveness of the proposed method.
	\end{itemize}
	
	\begin{figure*}[t]
		\centering
		\includegraphics[scale=0.51]{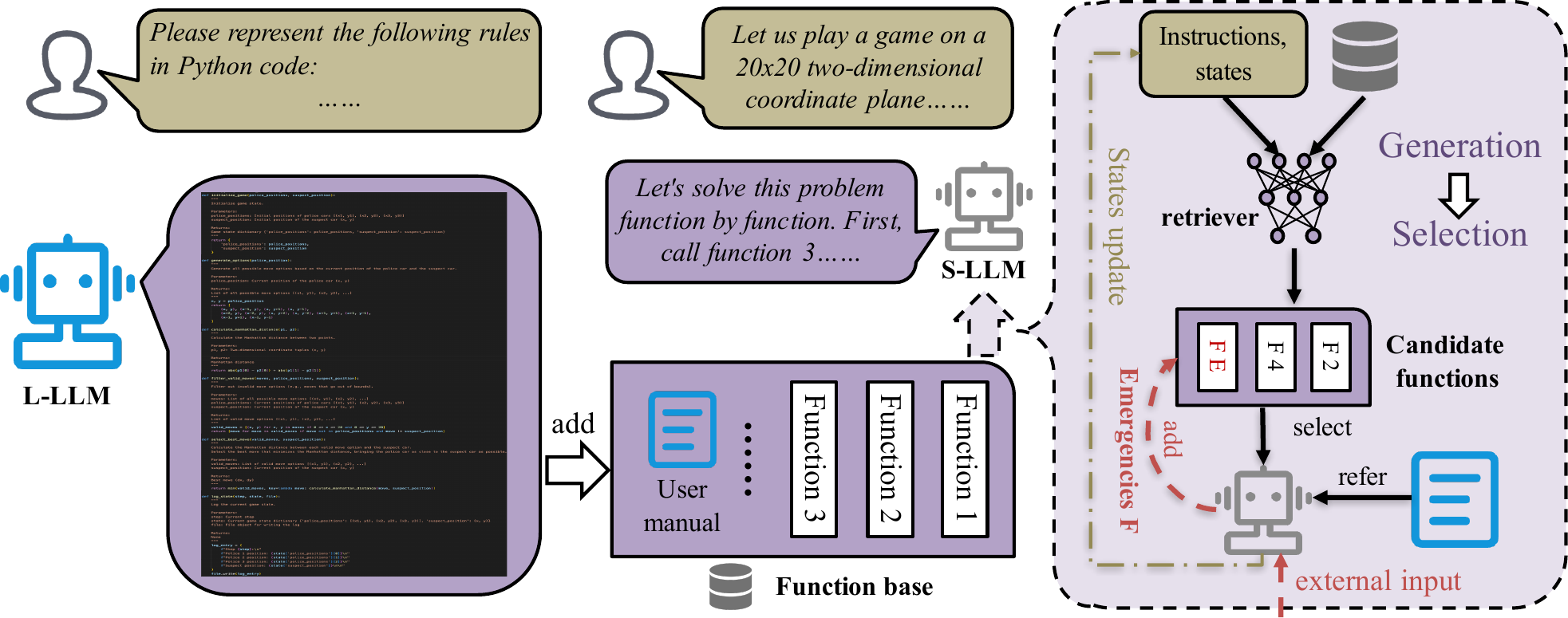}
		\caption{Illustration of the proposed Logic Distillation (LD). LD consists of three components: L-LLMs, retriever, and S-LLMs. L-LLMs are responsible for decomposing human-provided rules and instantiating them as basic functions to construct a function base. Besides, L-LLMs offer a user manual that explains the usage of these functions, including details such as rule descriptions and code comments.
			The retriever is in charge of retrieving the top-$K$ functions based on the insturctions and states. 
			S-LLMs select the appropriate functions for different stages of the task. Subsequently, S-LLMs will systematically make decisions function by function.}
		\label{Fig: LD}
	\end{figure*}
	
	\section{Related Work}
	Large language models (LLMs) \cite{ouyang2022training,chowdhery2022palm,thoppilan2022lamda,zhao2023survey,koundinya2024machine} are trained on broad data and can be easily adapted to a wide range of tasks \cite{bommasani2021opportunities}, which have been applied to education \cite{biswas2023role,kasneci2023chatgpt}, healthcare \cite{thirunavukarasu2023large,peng2023study,guo2024heart}, finance \cite{wu2023bloomberggpt}, etc. However, LLMs with impressive capabilities often suffer from size limitations, making them impractical to run on most devices and costly for invoking their interfaces \cite{chen2024data}. 

	In order to endow S-LLMs with the superior capabilities of L-LLMs, Knowledge Distillation (KD) of LLMs has become a focal point of related research \cite{xu2024survey}. In this context, L-LLMs like GPT-4 or GLM-4 are highly skilled teachers, while S-LLMs are students that learn to mimic the outputs of teachers \cite{he2023annollm,gu2024minillm,agarwal2024policy,liu2023llm}. Besides, numerous works have also focused on the reasoning steps of LLMs. Distilling step-by-step \cite{hsieh2023distilling} extracts L-LLM rationales as additional supervision for training S-LLMs within a multi-task framework. The Orca framework \cite{mukherjee2023orca} augments the prompt-response data pairs by incorporating a system message designed to facilitate student models' comprehension of the reasoning process. Subsequently, Orca 2 \cite{mitra2023orca} advances this approach by training the student model to discern the most efficacious solution strategy for individual tasks, guided by the performance metrics established by Orca. 
	However, these methods are essentially still making S-LLMs mimic the outputs of L-LLMs, rather than comprehending the reasoning logic.
	

	\section{Methodology}
	This paper investigates interactive decision-making tasks, where interactions can be divided into multiple steps (each interaction counts as one step), and each step can be further divided into multiple stages (several stages collectively complete one decision-making process).
	
	As illustrated in Figure \ref{Fig: LD}, we explore the Logic Distillation (LD) to empower S-LLMs to engage in decision-making akin to L-LLMs.
	LD first decomposes rules (instructions) $x$ into multiple stages with L-LLMs and converts the decision-making logic into the corresponding code functions (such as calculating the distance between points). Then, LD will construct a function base containing functions and user manual of these functions.
	Besides, LD employs a retriever to find the top-$K$ functions that most relevant to the current states and $x$. Subsequently, S-LLMs will select functions $f$ one by one to make decisions. 
	Overall, LD consists of three components: (i) L-LLMs, (ii) retriever, (iii) S-LLMs.
	
	\subsection{L-LLMs}
	L-LLMs exhibit exceptional capabilities in decision-making. To distill these capabilities into the S-LLMs, we propose instantiating the logic of L-LLMs through functions:
	\begin{equation}
	p_{\theta_L}(f, u|x)=\prod_{i}^{}p_{\theta_L}(y_i|x,y_{1:i-1})
	\label{Eq:1}
	\end{equation}
	where L-LLMs $p_{\theta_L}$ parametrized by $\theta_L$ that generates a current token $y_i$ based on a context of the previous $i -1$ tokens, multiple $y$ constitute a function $f$ and user manual (includes rule explanations, code comments, corresponding invocation stages, and so on). In addition, a collection of $f$ and $u$ forms the function base $D_f$.
	
	\subsection{Retriever}
	
	For the retriever $p_{\theta_R}(f|x,s)$, we have proposed two solutions tailored for function base of different scales. When the scale of the function base is large, we follow prior work \cite{lewis2020retrieval} to implement retrieval component based on DPR \cite{karpukhin2020dense}, and retriever $p_{\theta_R}(f|x,s)$ follows:
	\begin{equation}
	p_{\theta_R}(f|x,s)\propto exp (d(f)^{\top},q(x,s)) \rightarrow f_{c}=[f_1,\cdots, f_K]
	\label{Eq:2}
	\end{equation}
	where $s$ is current states, $d(f)$ is a dense representation of functions (including code comments and rule descriptions), and $q(x,s)$ is a query representation. Retriever $p_{\theta_R}(f|x,s)$ will return a top-$K$ list $f_{c}$, where the $K$ functions with highest prior probability.
	As for small-scale function base, we will fine-tune S-LLMs so that S-LLMs can directly select and utilize the appropriate functions from base $D_f$ to make decisions.

	\subsection{S-LLMs}
	We fine-tune S-LLMs to enable them to comprehend the functionality and the appropriate invocation timing of different functions. S-LLMs first select a function $f_j$ from $D_f$ or $f_{c}$ for stage $j$: 
	\begin{equation}
	\begin{aligned}
	&p_{\theta_S}(f_j|x,s) = max([p_{\theta_S}(f_1|x,s),\cdots,p_{\theta_S}(f_K|x,s)])\\ 
	\end{aligned}
	\label{Eq:3}
	\end{equation}
	
	Then, function $f_j$ will be executed to obtain the intermediate result of the $j$-th stage:
	\begin{equation}
	o_j=f_j(x,s),\quad s=o_{j-1}
	\label{Eq: stagej}
	\end{equation}
	
	If there are $J$ stages in a step of the task, S-LLMs will select $J$ functions, and the decision-making outputs for that step will be:
	\begin{equation}
	O=o_J=f_J(x,s),\quad s=o_{J-1}
	\label{Eq: stageJ}
	\end{equation}
	
	If $O$ meets the requirements of the task, the decision-making process will be halted. Otherwise, $O$ will be regarded as input for the next step.

	\subsection{Emergency handling of S-LLMs in LD}
	An advantage of LLMs is their ability to respond to various situations. When using LLMs to control embodied agents, they can respond to unforeseen circumstances. For instance, when LLMs control unmanned vessels for maritime exploration, they might navigate from the open sea to archipelagos, and LLMs can analyze the specific terrain, enabling swift traversal of the archipelago. 
	
	Typically, actions required in an emergency situation (such as avoiding whirlpools when controlling unmanned vessels) are simpler compared to the initial various instructions and rules. Therefore, in the LD framework, S-LLMs will transform emergency $x_E$ into functions and add them to the function candidate list $f_{c}$:
	\begin{equation}
	p_{\theta_S}(f_{E}, u|x_E,s)=\prod_{i}^{}p_{\theta_S}(y_i|x_E,s,y_{1:i-1}), f_E,u=\prod_{i}^{}y_i
	\label{Eq:E}
	\end{equation}
	Through Equation \ref{Eq:E}, S-LLMs will possess stronger general capabilities.
	
	
	It should be noted that, in contrast to KD, which necessitates S-LLMs to memorize massive L-LLMs' outputs, LD merely requires S-LLMs to remember the usage of functions. Consequently, LD preserves more general capabilities of LLMs, including function generation.
	The proposed LD is summarized in Algorithm \ref{A1}. 
	\begin{algorithm}[tb]
		\caption{Logic Distillation}
		\label{alg:train}
		\textbf{Input}: rules (instructions) $x$.\\
		\textbf{Parameter}: L-LLMs $p_{\theta_L}$, S-LLMs $p_{\theta_S}$, retriever $p_{\theta_R}$.\\
		\textbf{Output}: the decision-making outcome $[o_1,o_2,\cdots]$ .
		\begin{algorithmic}[1] 
			\STATE Generate functions $f$ and corresponding user manual $u$ with L-LLMs by Equation \ref{Eq:1}.
			\STATE Building function base $D_f$ with $f$ and $u$.
			\STATE  Initialize $O$, $s$.
			\WHILE {Decision-making output $O$ of one step does not meet the task requirements}
			\WHILE {$j$ in $1,2,\cdots,J$}
			\STATE Retrieve top-$K$ functions $[f_1,\cdots,f_K]$ with $p_{\theta_R}$, $x$ and $s$ by Equation \ref{Eq:2}.
			\STATE S-LLMs select the most suitable function $f_j$ from $[f_1,\cdots,f_K]$ for stage $j$.
			\STATE Obtaining intermediate results $o_j$ by Equation \ref{Eq: stagej}.
			\ENDWHILE
			\STATE $O,s=o_J$
			\IF {emergencies}
			\STATE Generate functions $f_E$ by Equation \ref{Eq:E} and add $f_E$ into $[f_1,\cdots,f_K]$.
			\ENDIF
			\ENDWHILE
		\end{algorithmic}
		\label{A1}
	\end{algorithm}
	
	\subsection{Why Selection Is Better }
	For the aforementioned limitation of S-LLMs, ineffectiveness in following complex instructions, we propose change the function of S-LLMs from generation to selection.
	Specifically, S-LLMs are required to select the appropriate functions from a provided set, which are to be employed at various stages when confronting a particular problem. This section theoretically analyzes the advantages of selection over generation.
	\begin{figure*}[t]
		\centering
		\includegraphics[scale=0.52]{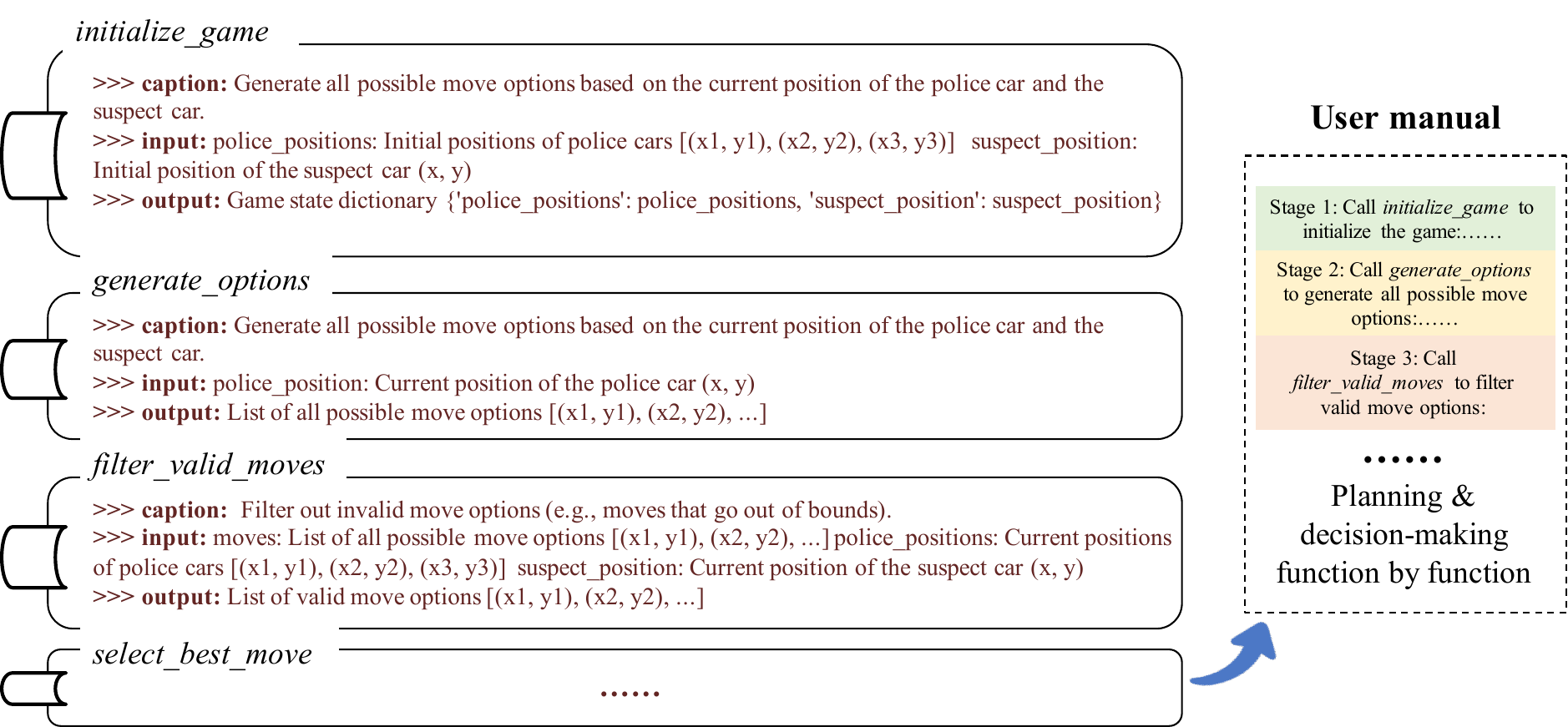}
		\caption{Function base. The L-LLM decomposes the rules and instantiates the decision-making logic into multiple functions (each function performs a specific task, such as calculating distance, etc.). In addition, the L-LLM enables the S-LLM to accurately invoke relevant functions by creating a user manual (including explanations, function comments, corresponding invocation stages, etc.), thereby completing the decision-making process.}
		\label{Fig: code}
	\end{figure*}
	
	Assuming that the token list of LLMs contains a total of $M$ types of tokens, the retriever provides $K$ types of functions. For generation, the entropy of the prediction is:
	\begin{equation}
	\begin{aligned}
	H_{generation} &=-\sum_{i=1}^M p^{'}_{\theta_S}(t_i) \log p^{'}_{\theta_S}(t_i),\\ \sum_{i=1}^M p^{'}_{\theta_S}(t_i)&=1
	\label{Eq:}
	\end{aligned}
	\end{equation}
	where $t_i$ is a token in the token list. 
	
	With Lagrange multiplier method \cite{liu1972method}, we get:
	\begin{equation}
	\begin{aligned}
	&Q\left(p^{'}_{\theta_S}(t_1), p^{'}_{\theta_S}(t_2), \ldots,p^{'}_{\theta_S}(t_M), \lambda\right)\\
	&=-\sum_{i=1}^M p^{'}_{\theta_S}(t_i) \log p^{'}_{\theta_S}(t_i)+\lambda\left(\sum_{i=1}^M p^{'}_{\theta_S}(t_i)-1\right)
	\label{Lagrange}
	\end{aligned}
	\end{equation}
	then partially differentiating $Q$ in Equation \ref{Lagrange} with respect
	to $p^{'}_{\theta_S}(t_i)$ and $\lambda$,
	\begin{align}
	\left\{
	\begin{aligned}
	&\frac{\partial Q}{\partial p^{'}_{\theta_S}(t_i)}=-\log p^{'}_{\theta_S}(t_i)-1+\lambda \\
	&\frac{\partial Q}{\partial \lambda}=\sum_{i=1}^M p^{'}_{\theta_S}(t_i)-1
	\end{aligned}
	\right.
	\label{KKT}
	\end{align}
	
	Let Equation \ref{KKT} be $0$, we can get:
	\begin{equation}
	\begin{aligned}
	&p^{'}_{\theta_S}(t_1)=p^{'}_{\theta_S}(t_2)=\ldots=p^{'}_{\theta_S}(t_M)=\frac{1}{M}, \\
	&H_{generation}=\log M
	\end{aligned}
	\end{equation}
	which is the maximum value of $H_{generation}$. 
	
	When we perform selection, the number of candidates will be $K$, and the maximum value of $H_{selection}$ will be $\log K$. As $K<<M$, $\log K << \log M$, the maximum value of $H_{selection}$ will be much smaller than that of $H_{generation}$, and the lower bound of selection will be much higher. Therefore, compared to generation, selection is more effective in maintaining stable outputs for S-LLMs.

	\begin{table*}[h]
		\centering
		\begin{tabular}{|c|c|c|c|c|}
			\hline
			\multicolumn{1}{|c|}{Methods}& Success & \multicolumn{1}{c|}{Failure without Violation} & \multicolumn{1}{c|}{Failure with Violation}& \multicolumn{1}{c|}{Average Steps of Success} \\
			\hline
			GLM4 (Large)        & 96.00\%  &  4.00\%  &  0.00\%   & 14.22 steps     \\
			GLM4-9B (Small)          &  0.00\%  &  0.00\%  & 100.00\%  & ---             \\
			GLM4-9B-KD       & 88.50\%  & 10.50\%  &  1.00\%   & 15.23 steps     \\
			GLM4-9B-LD       & \textbf{100.00\%} & \textbf{0.00\%} & \textbf{0.00\%} & \textbf{13.26 steps} \\
						\hline
			LLaMA3-70B (Large)       & 94.00\%  &  2.00\%  &  2.00\%   & 16.37 steps     \\
			LLaMA3-7B (Small)        &  0.00\%  &  2.00\%  & 98.00\%   & ---             \\
			LLaMA3-7B-KD     & 80.50\%  & 11.50\%  &  8.00\%   & 18.74 steps     \\
			LLaMA3-7B-LD     & \textbf{100.00\%} & \textbf{0.00\%} & \textbf{0.00\%} & \textbf{15.28 steps} \\
						\hline
			Qwen2.5-72B (Large)      & 95.50\%  &  4.00\%  &  0.50\%   & 13.92 steps     \\
			Qwen2.5-7B (Small)      &  1.00\%  &  0.00\%  & 99.00\%   & 56.85 steps     \\
			Qwen2.5-7B-KD    & 89.00\%  &  6.00\%  &  5.00\%   & 14.99 steps     \\
			Qwen2.5-7B-LD    & \textbf{100.00\%} & \textbf{0.00\%} & \textbf{0.00\%} & \textbf{13.23 steps} \\
			\hline
		\end{tabular}
			\caption{Results of pursuit game. 
		In this context, ``Success" refers to the successful conclusion of the game, where the blue dot captures the orange dot. ``Failure without Violation" indicates an unsuccessful outcome due to the inability to capture the orange dot within the specified number of moves. On the other hand, ``Failure with Violation" signifies an unsuccessful outcome resulting from a violation of the game rules. Lastly, ``Average Steps of Success" quantifies the average number of moves required for successful completion of the game. All methods are tested on 200 sets of starting positions.
	}
		\label{Tab:PG}
	\end{table*}
	
	\begin{figure*}[h]
		\centering	
		\subfigure[S-LLM with KD (Failure)]{
			\includegraphics[width=5.5 cm,height=4cm]{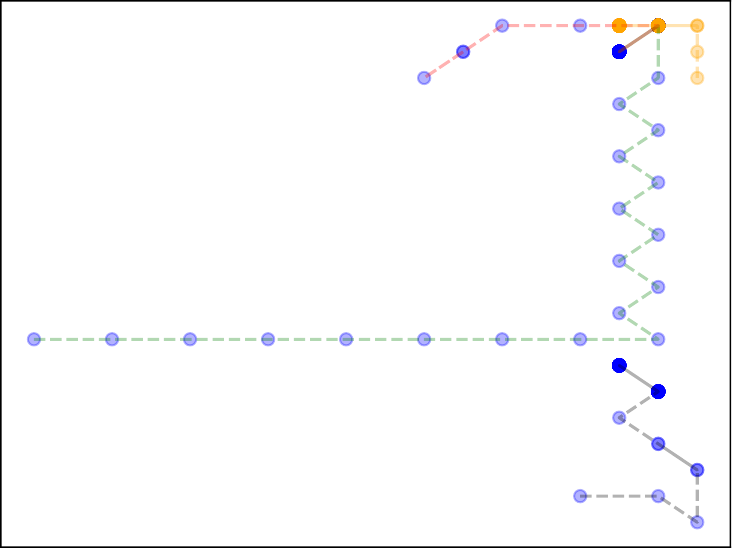}
			\label{Fig:A}
		}
		\subfigure[L-LLM (18 steps)]{
			\includegraphics[width=5.5 cm,height=4cm]{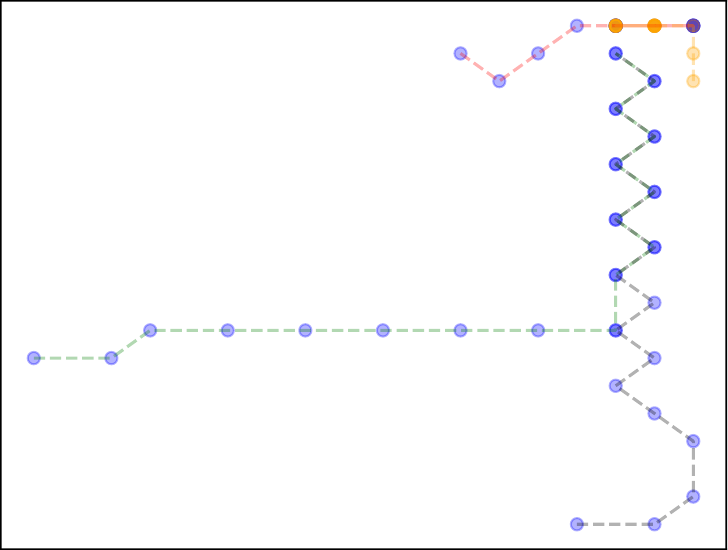}
			\label{Fig:B}
		}
		\subfigure[S-LLM with LD (15 steps)]{
			\includegraphics[width=5.5 cm,height=4cm]{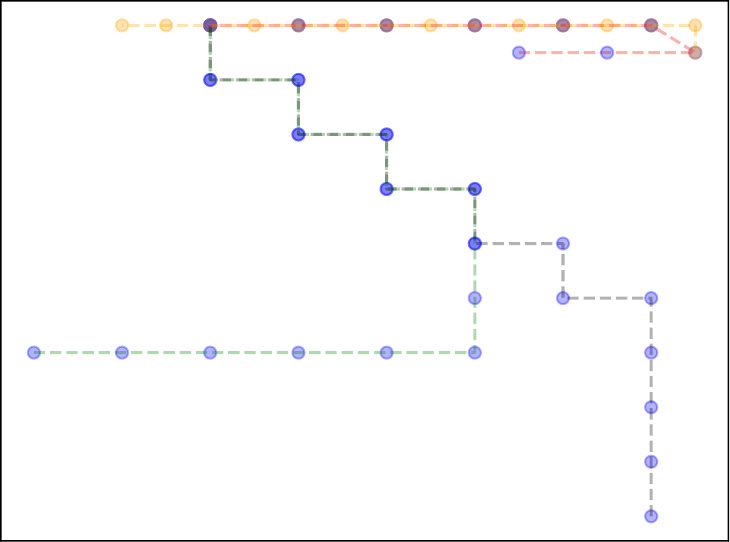}
			\label{Fig:C}
		}
		\\
		\caption{The global perspective of the pursuit game based on GLM4 and GLM4-9B. The trajectories of the three blue dots are represented by green, red, and grey dashed lines, respectively, with the darker colors indicating a higher number of passages.}
		\label{Fig: PG_overall}
	\end{figure*}

	\section{Experiments}
	Our experiments aim to: (1) verify the effectiveness of LD in decision-making tasks, (2) explore the reasons for the effectiveness of LD (3) verify that LD has a stronger ability to respond to emergencies.
	In our experiments, the L-LLMs are GLM-4, LLaMA3-70B and Qwen2.5-72B , while the S-LLMs are GLM4-9B, LLaMA3-7B and Qwen2.5-7B. 

	\subsection{Better Performance in Pursuit Game}
	We first conduct experiments based on the pursuit game to demonstrate that the proposed method can effectively enhance the decision-making capabilities of S-LLMs.
	More specifically, the pursuit game involves two sides, each controlled by a different LLM. One LLM manages three blue dots, while the other one controls an orange dot. Each interaction between the two sides constitutes a step. In each iteration, the blue dots are constrained to move by two units, while the orange dot is restricted to a single unit of movement. The game concludes when the Manhattan distance between all three blue dots and the orange dot is less than 2 units. 
	
	Specifically, the orange dot is consistently controlled by the original S-LLM, while the blue dots are managed by the L-LLM, S-LLM, S-LLM with KD, and S-LLM with LD, respectively. Additional selection and judgment rules have been introduced to improve the success rate of the baselines. For selection, LLMs are provided with the next decision coordinates to choose from. Regarding judgment, if LLMs make more than seven illegal choices, the game is considered a failure. The upper limit for the number of moves in the game is capped at 100. 
	To perform KD, we initialize 221 sets of starting positions randomly and produce 103,355 sets of outputs with the L-LLM. Subsequently, we fine-tune the S-LLM with LoRA \cite{hu2021lora} based on these outputs. 
	For LD, as depicted in Figure \ref{Fig: code}, we initially establish a function base with the L-LLM, where L-LLM decomposes the rules and instantiating the decision-making logic into multiple functions. 
	Moreover, to assist the S-LLM in learning how to utilize various functions and decide when to invoke them, L-LLM creates a user manual for these functions.
	By fine-tuning S-LLM with the user manual, it can comprehend the logic of the L-LLM and execute decision-making processes function by function.

	The results of the pursuit game are presented in Table \ref{Tab:PG}. It is evident that S-LLM struggles to comprehend complex instructions, as its failures mainly stem from rule violations.
	Conversely, the ``Failure with Violation" rates of L-LLMs are close to zero., demonstrating its superior comprehension and capability to follow instructions.
	By employing KD to mimic L-LLMs' outputs, S-LLMs' decision-making capabilities significantly improve. Nonetheless, their success rates are still notably lower than that of L-LLMs, and the number of steps in successful instances are higher.
	As for LD, the game’s success rates have reached $100\%$, and the average number of steps taken by S-LLMs is fewer than that of L-LLMs. These results comprehensively demonstrate the effectiveness of LD in enhancing S-LLMs' decision-making capabilities.

	\begin{table*}[t]
		\centering
		\begin{tabular}{|c|c|c|c|c|}
			\hline
			\multicolumn{1}{|c|}{Methods}& Success & \multicolumn{1}{c|}{Failure without Violation} & \multicolumn{1}{c|}{Failure with Violation}& \multicolumn{1}{c|}{Average Steps of Success} \\
			\hline
			GLM4 (Large)        & 90.00\%  &  7.50\%  &  2.50\%   & 18.98 steps      \\
			GLM4-9B (Small)          &  0.00\%  &  0.00\%  & 100.00\%  & ---              \\
			GLM4-9B-KD        & 52.00\%  &  2.50\%  & 45.50\%   & 18.5 steps       \\
			GLM4-9B-LD        & \textbf{100.00\%} & \textbf{0.00\%} & \textbf{0.00\%} & \textbf{14.65 steps} \\
			\hline
		\end{tabular}
			\caption{Results of pursuit game with emergencies. }
					\label{Tab:PGE}
	\end{table*}

	\subsection{Why LD is Better}
	
	In Figure \ref{Fig: PG_overall}, we initialize different LLMs from identical starting positions: blue dots at $(3,8), (14,19),(17,2)$, and the orange dot at $(20,18)$ to enable a global comparison of the overall decision-making capabilities among different LLMs.
	
	In Figure \ref{Fig:A}, as S-LLM with KD merely mimics the outputs of L-LLM, blue dots may represent outputs from the L-LLM in different scenarios, resulting in behaviors such as repetitive circling. For instance, the point along the grey trajectory continuously shuttle back and forth, making it impossible to catch up with the orange point.
	In Figure \ref{Fig:B}, the point controlled by L-LLM appears to backtrack, mainly because of the orange point’s continuous back-and-forth movements in an attempt to escape encirclement. 
	Besides, from Figure \ref{Fig:C}, it can be observed that S-LLM with LD enables the blue dots to approach the orange dot in a more direct manner. Thus, contrasting with L-LLM, S-LLM with LD requires fewer steps to successfully capture the target. Such results stem from different emphases: For LD, L-LLM employs a global perspective to design functions, causing the S-LLM to pay more attention to the overall distance from the orange dot. However, during one decision-making step, L-LLM is more susceptible to the influence of the current state, consequently neglecting the global planning.
	This is also why the performance of LD in Table \ref{Tab:PG} is superior to that of L-LLM.
	
	\subsection{Pursuit Game with Emergencies}
	In order to assess the capacity of different LLMs to handle emergencies, we introduced a $5\times5$ restricted area within the game plane. Related results based on GLM4 and GLM4-9B are presented in Table \ref{Tab:PGE}. 
	
	In the more intricate scenario, L-LLM can still grasp the rules through simple textual descriptions and implement effective decision-making, as the success rate achieve $90\%$. Conversely, S-LLM with KD achieves a success rate of 
	$52\%$, with failures resulting from rule violations (accounting for $45.5\%$), indicating an inability of S-LLM with KD to comprehend the new rules.
	As for S-LLM with LD, the process involves S-LLM initially abstracting the restricted area as a coordinate filtering function. Then, this function will handle the output of $filter\_valid\_moves$ from Figure \ref{Fig: code}, producing a list that excludes the coordinates of the restricted area. Subsequently, this list is utilized by $select\_best\_move$ to generate a suitable coordinate. The success rate of S-LLM with LD exceeds that of L-LLM by $10\%$, and its average number of steps is approximately $4$ steps fewer than the baselines, demonstrating the powerful general capabilities of LD.
	It should be noted that, compared to KD, LD only uses a few functional examples to fine-tune the S-LLM, enabling S-LLM to retain more general capabilities. 
	
\begin{figure}[t]
		\centering	
		\subfigure[L-LLM (22 steps)]{
			\includegraphics[width=4.05 cm,height=3.5cm]{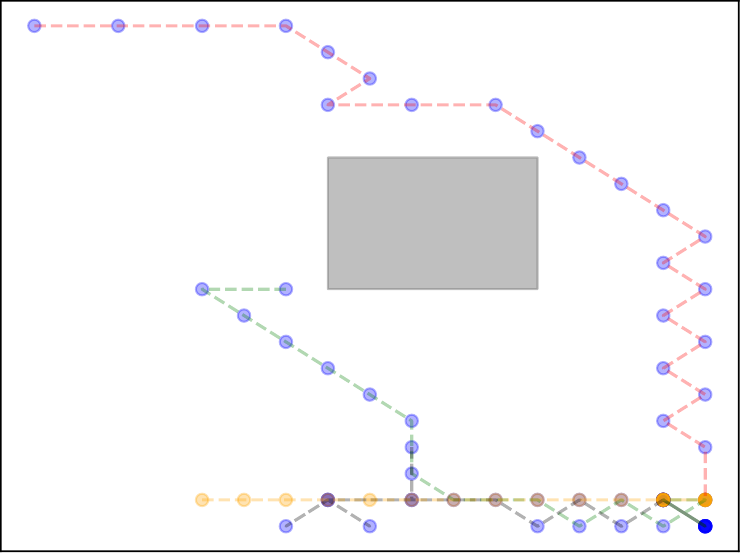}
			\label{Fig:PA}
		}
		\subfigure[S-LLM with LD (16 steps)]{
			\includegraphics[width=4.05 cm,height=3.5cm]{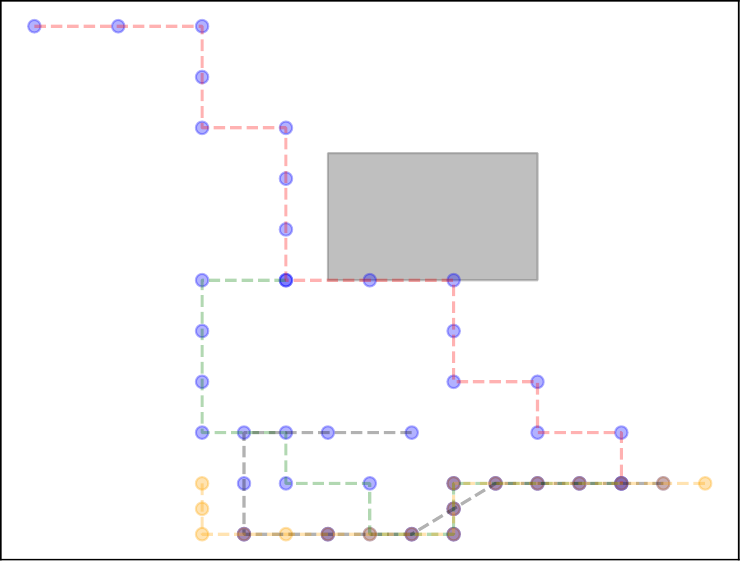}
			\label{Fig:PB}
		}
		\\
		\caption{The process of the pursuit game with emergencies.}
		\label{Fig: PGP_overall}
	\end{figure}

	In Figure \ref{Fig: PGP_overall}, we illustrate the comprehensive decision-making processes of L-LLM and S-LLM with LD, which further validates LD has a stronger ability to respond to emergencies. 
	
	This experiment also highlights the advantages of controlling embodied agents through LLMs. Traditional reinforcement learning methods for controlling embodied agents require retraining the model from scratch when encountering new influencing factors. In contrast, under the LD framework, S-LLMs can directly accept new rules and add them to the function library, enabling the agent to address new challenges by combining different functions.
	
	\subsection{Better Performance in 21 Ponits}

\begin{figure}[t]
		\centering
		\includegraphics[width=7cm]{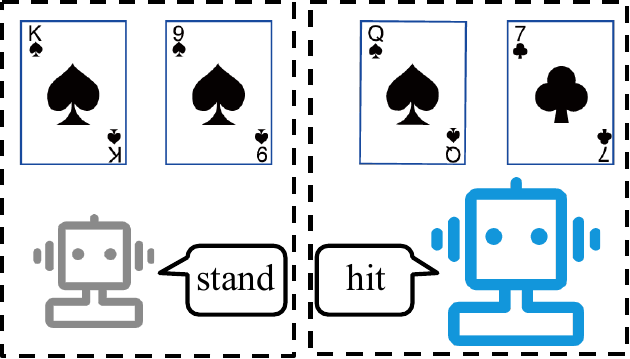}
		\caption{KD vs LD. KD aims to have smaller models mimic the output of larger models, while LD tries to enable smaller models to understand how larger models accomplish a task.}
		\label{Fig: 21}
	\end{figure}

	To further validate the improvement in S-LLM's decision-making capabilities with LD, we engage LLMs in a modified version of the game ``21 points." As depicted in Figure \ref{Fig: 21}, during each round, the two participating LLMs must decide whether to ``stand" or ``hit", with the objective of surpassing their opponent's total score without exceeding 21 points. Notably, cards ``10", ``J", ``Q", and ``K" are each assigned a value of 10, and card ``A" can be used as either 1 or 11. Furthermore, we streamline the game by reducing the deck to 26 cards of two suits, thereby enabling the LLMs to deduce the likelihood of exceeding 21 points after requesting a card, based on the available information. 

	\begin{table}[t]
		\centering
		\begin{tabular}{|c|c|c|c|}
			\hline
			\multicolumn{1}{|c|}{Methods}& Victory & \multicolumn{1}{c|}{Loss} & \multicolumn{1}{c|}{Draw}\\
			\hline
			GLM4 (Large)        & 55.50\%   & 42.00\%   & 2.50\%    \\
			GLM4-9B (Small)          & 36.50\%   & 62.00\%   & 1.50\%    \\
			GLM4-9B-KD        & 48.00\%   & 49.00\%   & 3.00\%    \\
			GLM4-9B-LD        & \textbf{59.50\%} & \textbf{38.50\%} & \textbf{2.00\%}  \\
			\hline
			LLaMA3-70B (Large)       & 57.00\%   & 42.00\%   & 1.00\%    \\
			LLaMA3-7B (Small)          & 39.50\%   & 60.00\%   & 0.50\%    \\
			LLaMA3-7B-KD      & 48.50\%   & 58.00\%   & 3.50\%    \\
			LLaMA3-7B-LD      & \textbf{58.50\%} & \textbf{38.50\%} & \textbf{3.00\%}  \\
			\hline
			Qwen2.5-72B (Large)      & 50.50\%   & 46.50\%   & 3.00\%    \\
			Qwen2.5-7B (Small)         & 34.00\%   & 65.50\%   & 0.50\%    \\
			Qwen2.5-7B-KD     & 44.50\%   & 54.50\%   & 0.00\%    \\
			Qwen2.5-7B-LD     & \textbf{55.00\%} & \textbf{43.00\%} & \textbf{2.00\%}  \\
			\hline
		\end{tabular}
			\caption{The victories, losses, and draws of each LLM in 200 times of 21 points. }
					\label{Tab:21}
	\end{table}
	
	Related results are presented in Table \ref{Tab:21}. 
It can be observed that L-LLMs demonstrate significant advantages in decision-making, as their win rates substantially exceed their rates of losses and draws.
	As for S-LLMs with KD, we first fine-tune S-LLMs with the outputs of L-LLMs. Although KD enhances S-LLMs' decision-making capabilities, the improvement is not significant, and there is still a gap between S-LLMs and L-LLMs. 
	In contrast, S-LLMs with LD achieve comparable or even superior results compaed to L-LLMs. 
	This is because L-LLMs teach the logic of decision-making to S-LLMs in the form of functions, enabling S-LLMs to maintain a global perspective. However, L-LLMs are susceptible to the influence of the current state in the decision-making process at each step, thereby losing its ability for global planning.
	
	\section{Conclusion}
	LLMs have been widely applied across many different fields. However, larger LLMs (L-LLMs) with powerful capabilities are difficult to deploy on the vast majority of devices due to their parameter scale. In contrast to L-LLMs, smaller open-source LLMs (S-LLMs) are easier to deploy but fall significantly short in performance compared to their larger counterparts.
	To improve the performance of S-LLMs, researchers have proposed various Knowledge Distillation (KD) methods. Nevertheless, KD merely enables S-LLMs to mimic the outputs of L-LLMs, which is insufficient for addressing decision-making problems. Thus, we propose Logic Distillation (LD), a method that instantiates the logic of L-LLMs by converting instructions into functions, thereby establishing a function base. Subsequently, through fine-tuning, S-LLMs will comprehend the usage of each function, enabling S-LLMs to make decisions. 
	Experimental results demonstrate that the proposed method, with a small amount of example fine-tuning, can enable S-LLMs to match or even surpass the decision-making capabilities of L-LLMs.

\section*{Acknowledgments}

This work is supported by National Natural Science Foundation of China (Grant Nos. 62436007, 62325602, 62036010, 62276238, U24A20326), the Foundation of Henan Educational Committee under Grant (No.25HASTIT034), the Natural Science Foundation of Henan under Grant (No.232300421095), the Key R\&D Projects in Zhejiang Province (No.2024C01106, 2025C01030), the Zhejiang NSF (LRG25F020001).

	\bibliographystyle{named}
	\bibliography{ijcai25}

\end{document}